\documentclass[11pt,a4paper]{article}
\usepackage[hyperref]{emnlp-ijcnlp-2019}
\usepackage{times}
\usepackage{latexsym}

\usepackage{times}
\usepackage{latexsym}
\usepackage{graphicx}
\usepackage{amsmath}
\usepackage{ amssymb }
\usepackage{booktabs}
\usepackage{multirow}

\usepackage{bbm}
\usepackage{bm}
\allowdisplaybreaks[4]

\usepackage{url}

\aclfinalcopy 

\title{Long and Diverse Text Generation with Planning-based Hierarchical Variational Model}

\author{
    Zhihong Shao\textsuperscript{1}, Minlie Huang\textsuperscript{1}\thanks{*Corresponding author: Minlie Huang.}, Jiangtao Wen\textsuperscript{1}, Wenfei Xu\textsuperscript{2}, Xiaoyan Zhu\textsuperscript{1}\\
    \textsuperscript{1} Institute for Artificial Intelligence, State Key Lab of Intelligent Technology and Systems \\
    \textsuperscript{1} Beijing National Research Center for Information Science and Technology \\
    \textsuperscript{1} Department of Computer Science and Technology, Tsinghua University, Beijing 100084, China \\
    \textsuperscript{2} Baozun, Shanghai, China\\
    {\tt szh19@mails.tsinghua.edu.cn, aihuang@tsinghua.edu.cn} \\
    {\tt jtwen@tsinghua.edu.cn, xuwenfeilittle@gmail.com, zxy-dcs@tsinghua.edu.cn}}

\date{}

\begin{document}

\maketitle
\begin{abstract}
Existing neural methods for data-to-text generation are still struggling to produce long and diverse texts: they are insufficient to model input data dynamically during generation, to capture inter-sentence coherence, or to generate diversified expressions. To address these issues, we propose a Planning-based Hierarchical Variational Model (PHVM). Our model first plans a sequence of groups (each group is a subset of input items to be covered by a sentence) and then realizes each sentence conditioned on the planning result and the previously generated context, thereby decomposing long text generation into dependent sentence generation sub-tasks. To capture expression diversity, we devise a hierarchical latent structure where a global planning latent variable models the diversity of reasonable planning and a sequence of local latent variables controls sentence realization. Experiments show that our model outperforms state-of-the-art baselines in long and diverse text generation.
\end{abstract}

\section{Introduction}
Data-to-text generation is to generate natural language texts from structured data~\cite{DBLP:journals/jair/GattK18}, which has a wide range of applications (for weather forecast, game report, product description, advertising document, etc.). Most neural methods focus on devising encoding scheme and attention mechanism, namely, (1) exploiting input structure to learn better representation of input data~\cite{DBLP:conf/emnlp/LebretGA16,DBLP:conf/aaai/LiuWSCS18}, and (2) devising attention mechanisms to better employ input data~\cite{DBLP:conf/naacl/MeiBW16,DBLP:conf/aaai/LiuWSCS18,DBLP:conf/naacl/NemaSJLSK18} or to dynamically trace which part of input has been covered in generation~\cite{DBLP:conf/emnlp/KiddonZC16}. These models are able to produce fluent and coherent short texts in some applications.

However, to generate long and diverse texts such as product descriptions, existing methods are still unable to capture the complex semantic structures and diversified surface forms of long texts. \textbf{First}, existing methods are not good at modeling input data dynamically during generation. Some neural methods~\cite{DBLP:conf/emnlp/KiddonZC16,DBLP:conf/ijcai/FengLL0SL18} propose to record the accumulated attention devoted to each input item. However, these records may accumulate errors in representing the state of already generated prefix, thus leading to wrong new attention weights. \textbf{Second}, inter-sentence coherence in long text generation is not well captured~\cite{DBLP:conf/emnlp/WisemanSR17} due to the lack of high-level planning. Recent studies propose to model planning but still have much space for improvement. For instance, in \cite{DBLP:journals/corr/abs-1809-00582} and \cite{DBLP:conf/aaai/ShaMLPLCS18}, planning is merely designed for ordering input items, which is limited to 
aligning input data with the text to be generated. 
\textbf{Third}, most methods fail to generate diversified expressions. Existing data-to-text methods inject variations at the conditional output distribution, which is proved to capture only low-level variations of expressions~\cite{DBLP:conf/aaai/SerbanSLCPCB17}.

\begin{figure}[!t]
    \includegraphics[width=0.5\textwidth]{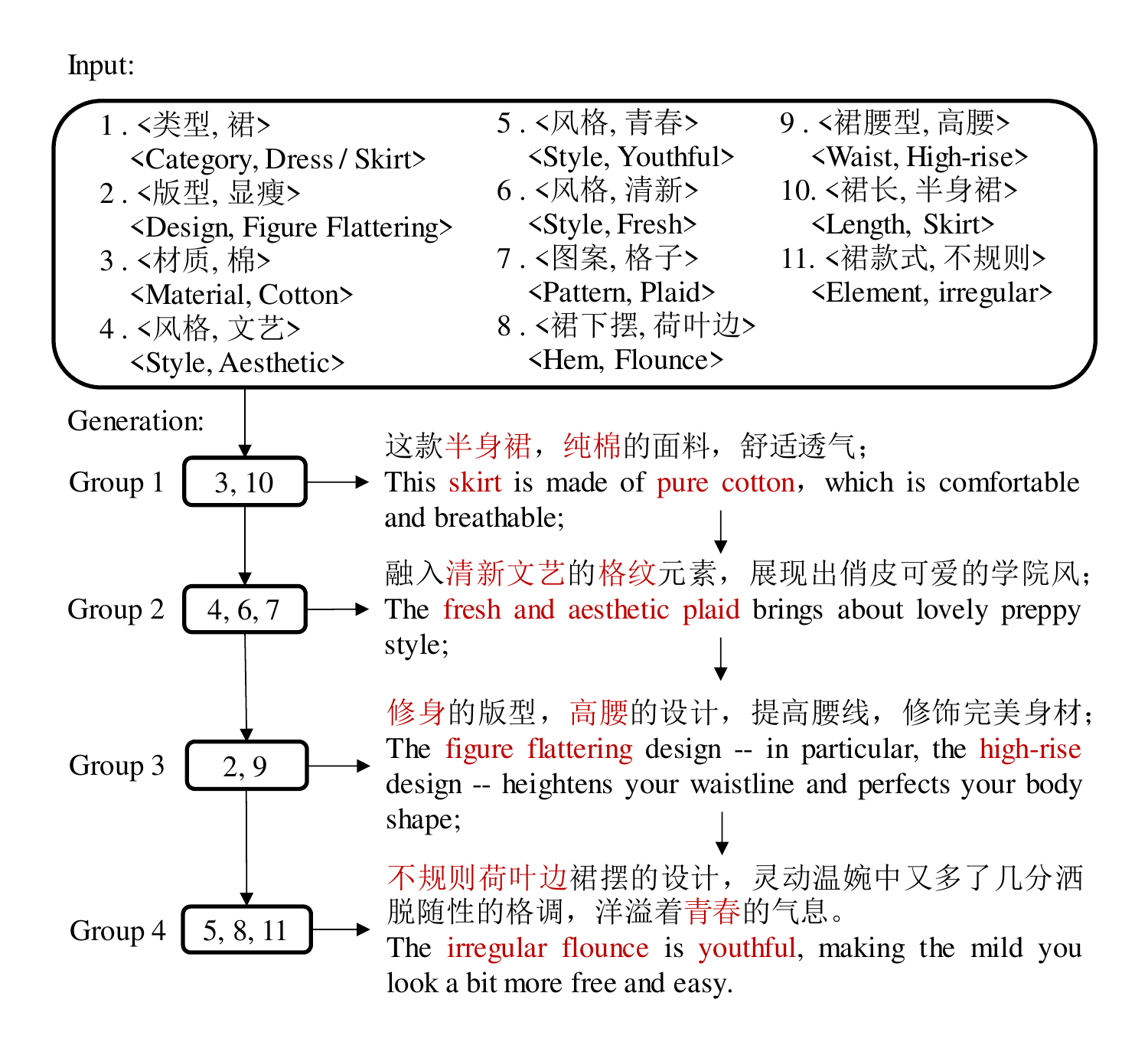}
    \caption{Generation process of PHVM. After encoding a list of input attribute-value pairs, PHVM first conducts planning by generating a sequence of groups, each of which is a subset of input items. Each sentence is then realized conditioned on the corresponding group and its previous generated sentences. }
    \label{fig:example} 
\end{figure}

To address the above issues, we propose a novel Planning-based Hierarchical Variational Model (PHVM). 
To better model input data and alleviate the inter-sentence incoherence problem, we design a novel planning mechanism and adopt a compatible hierarchical generation process, 
which mimics the process of human writing. 
Generally speaking, to write a long text, a human writer first arranges contents 
and discourse structure 
(i.e., {\it high-level planning}) and then realizes the surface form of each individual part ({\it low-level realization}). 
Motivated by this, our proposed model first performs planning by segmenting input data into a sequence of groups, 
and then generates a sentence conditioned on the corresponding group and preceding generated sentences. 
In this way, we decompose long text generation into a sequence of dependent sentence generation sub-tasks where each sub-task depends specifically on an individual group and the previous context. By this means, the input data can be well modeled and inter-sentence coherence can be captured. Figure \ref{fig:example}
depicts the process.

To deal with expression diversity, this model also enables us to inject variations at both high-level planning and low-level realization with a hierarchical latent structure. At high level, we introduce a global planning latent variable to model the diversity of reasonable planning. 
At low level, we introduce local latent variables for sentence realization. Since our model is based on Conditional Variational Auto-Encoder (CVAE) \cite{DBLP:conf/nips/SohnLY15}, expression diversity can be captured by the global and local latent variables. 

We evaluate the model on a new advertising text\footnote{An advertising text describes a product with attractive wording. The goal of writing such texts is to advertise a product and attract users to buy it.} 
generation task which requires the system to generate a long and diverse advertising text that covers a given set of attribute-value pairs describing a product (see Figure \ref{fig:example}). 
We also evaluate our model on the recipe text generation task from \cite{DBLP:conf/emnlp/KiddonZC16} which requires the system to correctly use the given ingredients and maintain coherence among cooking steps. 
Experiments on advertising text generation show that our model outperforms state-of-the-art baselines in automatic and manual evaluation. Our model also generalizes well to long recipe text generation and outperforms the baselines.
Our contributions are two-fold:
\begin{itemize}
    \item We design a novel Planning-based Hierarchical Variational Model (PHVM) which integrates planning into a hierarchical latent structure. Experiments show its effectiveness in coverage, coherence, and diversity.
    
    \item We propose a novel planning mechanism which segments the input data into a sequence of groups, thereby decomposing long text generation into dependent sentence generation sub-tasks. Thus, input data can be better modeled and inter-sentence coherence can be better captured.
    To capture expression diversity, we devise a hierarchical latent structure which injects variations at both high-level planning and low-level realization. 
\end{itemize}

\begin{figure*}[htb!]
    \centering
    \includegraphics[width=\textwidth]{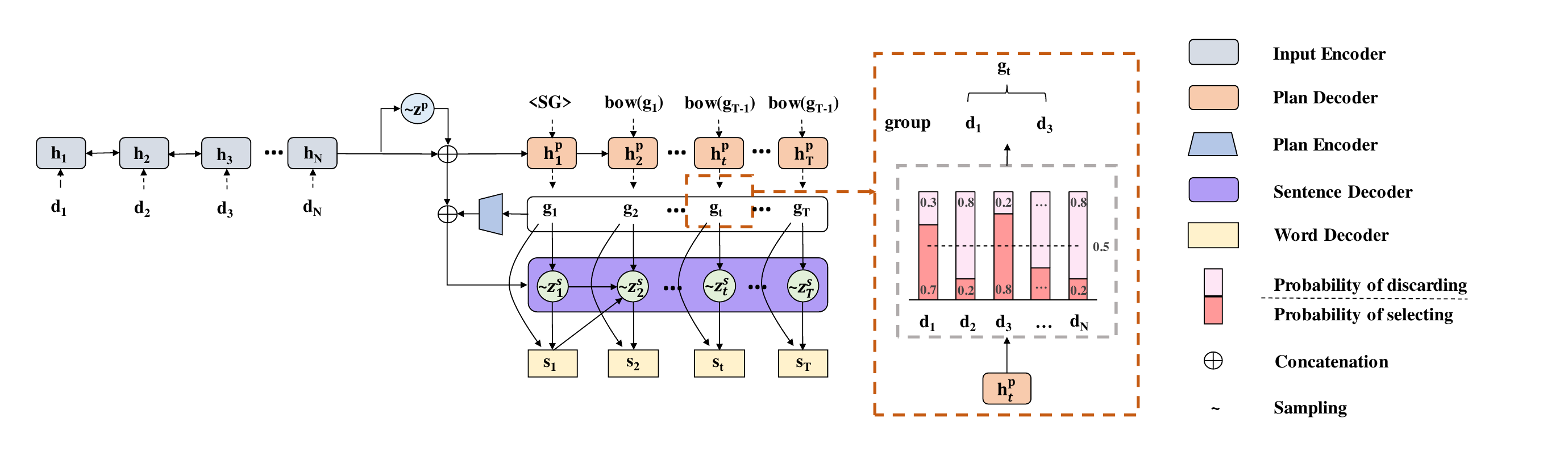}
    \caption{Architecture of PHVM. The model controls planning with a global latent variable $z^p$. The plan decoder conducts planning by generating a sequence of groups $g=g_1g_2...g_T$ where $g_t$ is a subset of input items and specifies the content of sentence $s_t$ to be generated. The sentence decoder controls the realization of $s_t$ with a local latent variable $z_t^s$; dependencies among $z_t^s$ are explicitly modeled to better capture inter-sentence coherence.
    }
    \label{fig:architecture}
\end{figure*}

\section{Related Work}
Traditional methods~\cite{DBLP:journals/nle/ReiterD97,DBLP:conf/acl/StentPW04} for data-to-text generation consist of three components: content planning, sentence planning, and surface realization. Content planning and sentence planning are responsible for what to say and how to say respectively; they are typically based on hand-crafted ~\cite{DBLP:conf/acl/Kukich83,DBLP:conf/ewnlg/DalianisH93,DBLP:journals/ai/Hovy93} or automatically-learnt rules ~\cite{duboue2003statistical}. Surface realization generates natural language by carrying out the plan, which is template-based \cite{mcroy2003augmented,DBLP:journals/coling/DeemterTK05} or grammar-based \cite{DBLP:journals/nle/Bateman97,DBLP:conf/acl/EspinosaWM08}. As these models are shallow and the two stages (planning and realization) often function separately, traditional methods are unable to capture rich variations of texts. 

Recently, neural methods have become the mainstream models for data-to-text generation due to their strong ability of representation learning and scalability. These methods perform well in generating weather forecasts~\cite{DBLP:conf/naacl/MeiBW16} or very short biographies~\cite{DBLP:conf/emnlp/LebretGA16,DBLP:conf/aaai/LiuWSCS18,DBLP:conf/aaai/ShaMLPLCS18,DBLP:conf/naacl/NemaSJLSK18} using well-designed data encoder and attention mechanisms. However, as demonstrated in  \newcite{DBLP:conf/emnlp/WisemanSR17} (a game report generation task), existing neural methods are still problematic for long text generation: they often generate incoherent texts. In fact, these methods also lack the ability to model diversity of expressions.

As for long text generation, recent studies tackle the incoherence problem from different perspectives. To keep the decoder aware of the crucial information in the already generated prefix, \newcite{DBLP:conf/emnlp/ShaoGBGSK17} appended the generated prefix to the encoder, and \newcite{DBLP:conf/aaai/GuoLCZYW18} leaked the extracted features of the generated prefix from the discriminator to the generator in a Generative Adversarial Nets~\cite{DBLP:conf/nips/GoodfellowPMXWOCB14}. To model dependencies among sentences, \newcite{DBLP:conf/acl/LiLJ15} utilized a hierarchical recurrent neural network (RNN) decoder. \newcite{DBLP:conf/emnlp/KonstasL13} proposed to plan content organization with grammar rules while \newcite{DBLP:journals/corr/abs-1809-00582} planned by reordering input data. Most recently, \newcite{DBLP:journals/corr/abs-1904-03396} proposed to select plans from all possible ones, which is infeasible for large inputs.

As for diverse text generation, existing methods can be divided into three categories: enriching conditions~\cite{AAAI1714563}, post-processing with beam search and rerank~\cite{DBLP:conf/naacl/LiGBGD16}, and designing effective models~\cite{D18-1428}. Some text-to-text generation models \cite{DBLP:conf/aaai/SerbanSLCPCB17,DBLP:conf/acl/ZhaoZE17} inject high-level variations with latent variables. Variational Hierarchical Conversation RNN (VHCR) \cite{DBLP:conf/naacl/ParkCK18} is a most similar model to ours, which also adopts a hierarchical latent structure. Our method differs from VHCR in two aspects: (1) VHCR has no planning mechanism, and the global latent variable is mainly designed to address the KL collapse problem, while our global latent variable captures the diversity of reasonable planning; (2) VHCR injects distinct local latent variables without direct dependencies, while our method explicitly models the dependencies among local latent variables to better capture inter-sentence connections. \newcite{DBLP:conf/acl/ShenCZCWGC19} proposed ml-VAE-D with multi-level latent variables. However, the latent structure of ml-VAE-D consists of two global latent variables: the top-level latent variable is introduced to learn a more flexible prior of the bottom-level latent variable which is then used to decode a whole paragraph. By contrast, our hierarchical latent structure is tailored to our planning mechanism: the top level latent variable controls planning results and a sequence of local latent variables is introduced to obtain fine-grained control of sentence generation sub-tasks.

We evaluated our model on a new advertising text generation task which is to generate a long and diverse text that covers all given specifications about a product. Different from our task, the advertising text generation task in \cite{DBLP:conf/kdd/ChenLZYZ019} is to generate personalized product description based on product title, product aspect (e.g., ``appearance"), and user category.

\section{Task Definition}
Given input data $x = \{d_1, d_2, ..., d_N\}$ where each $d_i$ can be an attribute-value pair or a keyword, our task is to generate a long and diverse text $y = s_1 s_2 ...s_T$ ($s_t$ is the $t^{th}$ sentence) that covers $x$ as much as possible. For the advertising text generation task, $x$ consists of specifications about a product where each $d_i$ is an attribute-value pair $<a_i, v_i>$. For the recipe text generation task, $x$ is an ingredient list where each $d_i$ is an ingredient. Since the recipe title $r$ is also used for generation, we abuse the symbol $x$ to represent $<\{d_1, d_2, ..., d_N\}, r>$ for simplification.

\section{Approach}
\subsection{Overview}
Figure \ref{fig:architecture} shows the architecture of PHVM. PHVM first samples a global planning latent variable $z^p$ based on the encoded input data; $z^p$ 
serves as a condition variable in both planning and hierarchical generation process. The plan decoder takes $z^p$ as initial input. At time step $t$, it decodes a group $g_t$ which is a subset of input items ($d_i$) and specifies the content of the $t^{th}$ sentence $s_t$. When the plan decoder finishes planning, the hierarchical generation process starts, which involves the high-level sentence decoder and the low-level word decoder. The sentence decoder models inter-sentence coherence in semantic space by computing a sentence representation $h_t^s$ and sampling a local latent variable $z_t^s$ for each group. $h_t^s$ and $z_t^s$, along with $g_t$, guide the word decoder to realize the corresponding sentence $s_t$.

The planning process decomposes the long text generation task into a sequence of dependent sentence generation sub-tasks, thus facilitating the hierarchical generation process. With the hierarchical latent structure, PHVM is able to capture multi-level variations of texts.

\subsection{Input Encoding}
We first embed each input item $d_i$ into vector $\bm{e}(d_i)$. The recipe title $r$ is also embedded as $\bm{e}(r)$. We then encode $x$\footnote{For advertising text generation, $x$ is first ordered by attributes so that general attributes are ahead of specific ones; for recipe text generation, we retain the order in the dataset} with a bidirectional Gated Recurrent Unit (GRU)~\cite{DBLP:conf/ssst/ChoMBB14}. For advertising text generation, $x$ is represented as the concatenation of the last hidden states of the forward and backward GRU $\bm{enc}(x) = [\overrightarrow{h_N};\overleftarrow{h_1}]$; for recipe text generation, $\bm{enc}(x) = [\overrightarrow{h_N};\overleftarrow{h_1};\bm{e}(r)]$. $h_i = [\overrightarrow{h_i};\overleftarrow{h_i}]$ is the context-aware representation of $d_i$. Note that input encoder is not necessarily an RNN; other neural encoders or even other encoding schemes are also feasible, such as multi-layer perceptron (MLP) and bag of words.

\subsection{Planning Process}
The planning process generates a subset of input items to be covered for each sentence, thus decomposing long text generation into easier dependent sentence generation sub-tasks. 
Due to the flexibility of language, there may exist more than one reasonable 
text that covers the same input but in different order. 
To capture such variety, we model the diversity of reasonable planning with a global planning latent variable $z^p$. 
Different samples of $z^p$ may lead to different planning results 
which control the order of content.
This process can be formulated as follows:
\begin{align}
	g &=argmax_{g} P(g|x, z^p) \label{eq:plan}
\end{align}
where $g= g_1g_2...g_T$ is a sequence of groups, and each group $g_t$ is a subset of input items which is a main condition when realizing the sentence $s_{t}$. 

The global latent variable $z^p$ is assumed to follow the isotropic Gaussian distribution, and is sampled from its prior distribution $p_{\theta}(z^p|x) = \mathcal{N}(\mu^p, \sigma^{p2}\textbf{I})$ during inference and from its approximate posterior distribution $q_{\theta^{'}}(z^p|x, y) = \mathcal{N}(\mu^{p'}, \sigma^{p'2}\textbf{I})$ during training:
\begin{align}
	[\mu^p; \log \sigma^{p2}] &= \bm{MLP}_{\theta}(x)\\
	[\mu^{p'}; \log \sigma^{p'2}] &= \bm{MLP}_{\theta^{'}}(x, y)
\end{align}

We solve Eq. \ref{eq:plan} greedily by computing $g_t = argmax_{g_t}P(g_t|g_{<t}, x, z^p)$ with the plan decoder (a GRU). Specifically, at time step $t$, the plan decoder makes a binary prediction for each input item by estimating $P(d_i \in g_t|g_{<t}, x, z^p)$: 

\begin{equation}
	P(d_i \in g_t) = \bm{\sigma}(v_p^T \bm{tanh}(W_p [h_i; h_t^p] + b_p))
\end{equation}
where $\bm{\sigma}$ denotes the sigmoid function, 
$h_i$ is the vector of input item $d_i$, 
and $h_t^p$ is the hidden state of the plan decoder. 
Each group is therefore formed as $g_t = \{d_i | P(d_i \in g_t) > 0.5\}$ (If this is empty, we set $g_t$ as $ \{argmax_{d_i} P(d_i \in g_t)\}$.). 

We feed $bow(g_t)$ (the average pooling of $\{h_i|d_i \in g_t\}$) to the plan decoder at the next time step, so that $h_{t+1}^p$ is aware of what data has been selected and what has not. The planning process proceeds until the probability of stopping at the next time step is over 0.5:
\begin{equation}
	P_t^{stop} = \bm{\sigma}(W_c h_t^p + b_c)
\end{equation}

The hidden state is initialized with $\bm{enc}(x)$ and $z^p$. 
The plan decoder is trained with full supervision, which is applicable to those tasks where reference plans are available or can be approximated. For both tasks we evaluate in this paper, we approximate the reference plans by recognizing the subset of input items covered by each sentence with string match heuristics. 
The loss function at time step $t$ is given by:
\begin{equation}
	\begin{split}
		-\log P(g_t&=\widetilde{g_t}|\widetilde{g_{<t}}, x, z^p) \\
		= &- \sum\limits_{d_i \in \widetilde{g_t}} \log P(d_i \in g_t) \\
		&- \sum\limits_{d_i \notin \widetilde{g_t}} \log (1 - P(d_i \in g_t))
	\end{split}
\end{equation}
where $\widetilde{g_t}$ is the reference group. As a result, $z^p$ is forced to capture features of reasonable planning. 

\subsection{Hierarchical Generation Process}
The generation process produces a long text $y=s_1s_2 ...s_T$ in alignment with the planning result $g=g_1 g_2 ...g_T$, which is formulated as follows:
\begin{align}
	c &= \{x, z^p\}\\
	y &= argmax_{y} P(y|g, c) \label{eq:gen}
\end{align}

We perform sentence-by-sentence generation and solve Eq. \ref{eq:gen} greedily by computing $s_t = argmax_{s_t} P(s_t|s_{<t}, g, c)$. $s_t$ focuses more on $g_t$ than on the entire plan $g$. The generation process is conducted hierarchically, which consists of sentence-level generation and word-level generation. Sentence-level generation models inter-sentence dependencies at high level, and interactively controls word-level generation which conducts low-level sentence realization. 

\paragraph{Sentence-level Generation}
The sentence decoder (a GRU) performs sentence-level generation; for each sentence $s_t$ to be generated, it produces a sentence representation $h_t^s$ and introduces a local latent variable $z_{t}^s$ 
 to control sentence realization.

The latent variable $z_t^s$ is assumed to follow the isotropic Gaussian distribution. At time step $t$, the sentence decoder samples $z_t^s$ from the prior distribution $p_{\phi}(z_t^s|s_{<t}, g, c) = \mathcal{N}(\mu_t^s, \sigma_t^{s2}\textbf{I})$ during inference and from the approximate posterior distribution $q_{\phi^{'}}(z_t^s|s_{\leq t}, g, c) = \mathcal{N}(\mu_t^{s'}, \sigma_t^{s'2}\textbf{I})$ during training. $h_t^s$ and the distribution of $z_t^s$ are given by:
\begin{align}
    h_t^s = \bm{GRU}&_s([z_{t-1}^s;h_{t-1}^w], h_{t-1}^s)\\
    [\mu_t^s;\log \sigma_t^{s2}] &= \bm{MLP}_{\phi}(h_t^s, bow(g_t))\\
	[\mu_t^{s'};\log \sigma_t^{s'2}] =& \bm{MLP}_{\phi^{'}}(h_t^s, bow(g_t), s_t)
\end{align}
where $h_{t-1}^w$ is the last hidden state of the word decoder after decoding sentence $s_{t-1}$, and $\bm{GRU}_s$ denotes the GRU unit of the sentence decoder. By this means, we constrain the distribution of $z_t^s$ in two aspects. 
First, to strengthen the connection from the planning result $g$, we additionally condition $z_t^s$ on $g_t$ to keep $z_t^s$ focused on $g_t$. 
Second, to capture the dependencies on $s_{<t}$, we explicitly model the dependencies among local latent variables by inputting $z_{t-1}^s$ to the sentence decoder, so that $z_t^s$ is conditioned on $z_{<t}^s$ and is expected to model smooth transitions in a long text. 

We initialize  the hidden state $h_0^s$ by encoding the input $x$, the global planning latent variable $z^p$ and the planning result $g$:
\begin{align}
    h_t^g &= \bm{GRU}_g(bow(g_t), h_{t-1}^g)\\
    h_0^s &= W_s [\bm{enc}(x);z^p;h_T^g] + b_s
\end{align}
where $h_T^g$ is the last hidden state of $\bm{GRU}_g$ that encodes the planning result $g$. 

\paragraph{Word-level Generation}
The word decoder (a GRU) conducts word-level generation; it decodes a sentence $s_t = argmax_{s_t} P(s_t|s_{<t}, z_t^s, g, c)$ conditioned on $\{h_t^s, z_t^s, g_t\}$. Specifically, we sample word $w^t_k$ of $s_t$ as follows: 
\begin{equation}
	w^t_k \sim P(w^t_k|w_{<k}^t, s_{<t}, z_t^s, g, c)
\end{equation}

\subsection{Loss Function}
We train our model end-to-end.
The loss function has three terms: the negative evidence lower bound (ELBO) of $\log P(y|x)$ ($\mathcal{L}_1$), the loss of predicting the stop signal ($\mathcal{L}_2$) and the bag-of-word loss~\cite{DBLP:conf/acl/ZhaoZE17} ($\mathcal{L}_3$).

We first derive the ELBO: 
\begin{gather}
	\begin{aligned}
	\log P(y|x) &\geq E_{q_{\theta^{'}}(z^p|x, y)}[\log P(y|x, z^p)]\\
	&-D_{KL}(q_{\theta^{'}}(z^p|x,y)||p_{\theta}(z^p|x))
	\end{aligned}\label{eq:e1}\\
	\begin{aligned}
	\log P(y|x, z^p) &= \log P(g, y|x, z^p) \\
                    &= \sum_{t=1}^T\log P(g_t|g_{<t}, x, z^p)\\
				    &+\log P(s_t|s_{<t}, g, x, z^p)
	\end{aligned}\label{eq:e2}\\
	\begin{aligned}
	&\log P(s_t|s_{<t}, g, x, z^p) \\
		&\geq  E_{q_{\phi^{'}}(z_t^s|s_{\leq t}, g, x, z^p)} [\log P(s_t|s_{<t}, z_t^s, g, x, z^p)] \\
		&- D_{KL}(q_{\phi^{'}}(z_t^s|s_{\leq t}, g, x, z^p)||p_{\phi}(z_t^s|s_{<t}, g, x, z^p))
	\end{aligned}\label{eq:e3}
\end{gather}
We can obtain the ELBO by unfolding the right hand side of Eq. \ref{eq:e1} with Eq. \ref{eq:e2} and \ref{eq:e3}. During training, we use linear KL annealing technique to alleviate the KL collapse problem\cite{DBLP:conf/conll/BowmanVVDJB16}.

$\mathcal{L}_2$ is given by:
\begin{equation}
	\mathcal{L}_2 = \sum_{t=1}^{T-1} \log P_t^{stop} + \log (1 - P_T^{stop})
\end{equation}

$\mathcal{L}_3$ is the sum of bag-of-word loss~\cite{DBLP:conf/acl/ZhaoZE17} applied to each sentence, which is another technique to tackle the KL collapse problem.

\section{Experiments}
\subsection{Dataset}
We evaluated PHVM on two generation tasks. The first task is the new advertising text generation task which is to generate a long advertising text that covers all given attribute-value pairs for a piece of clothing. The second task is the recipe generation task from \cite{DBLP:conf/emnlp/KiddonZC16} which is to generate a correct recipe for the given recipe title and ingredient list.

\paragraph{Advertising Text Generation}
We constructed our dataset from a Chinese e-commerce platform. 
The dataset consists of 119K pairs of advertising text and clothing specification table. 
Each table is a set of attribute-value pairs describing a piece of clothing. 
We made some modifications to the original specification tables.  Specifically, if some attribute-value pairs from a table do not occur in the corresponding text, the pairs are removed from the table.  We also recognized attribute values by string matching with a dictionary of attribute values. If a pair occurs in the text but not in the table, the pair is added to the table. 

The statistics are shown in Table \ref{tab:dstat} and Table \ref{tab:gstat}.

\begin{table}[htbp!]
	\centering
	\small
	\begin{tabular}{*{4}{c}}
		\toprule
		Category & Tops & Dress / Skirt & Pants \\
		\midrule
		\# Type & 22 & 23 & 9 \\
		\# Attr. & 13 & 16 & 11 \\
		\# Val. & 264 & 284 & 203\\
		Avg. \# Input Pairs & 7.7 & 7.7 & 6.6 \\
		Avg. Len. & 110 & 111 & 108\\
		\# Instances & 48K & 47K & 24K\\
		\bottomrule
	\end{tabular}
	\caption{\label{tab:dstat}Detailed statistics of our dataset. \textit{\# Attr.} / \textit{\# Val.}: the total number of attributes / attribute values.}
\end{table}
\begin{table}[htbp!]
\centering
\resizebox{0.5\textwidth}{!}{
\begin{tabular}{*{5}{c}}
  \toprule
  \# Attr. & \# Val. & Vocab & Avg. \# Input Pairs & Avg. \# Len.\\
  \midrule
   28 & 633 & 54.9K & 7.5 & 110.2\\
  \bottomrule
\end{tabular}}
\caption{\label{tab:gstat}General statistics of our dataset. We counted the size of vocabulary after removing brand names.}
\end{table}
\begin{table*}[!t]
    \centering
	\small
    \begin{tabular}{*{6}{c}}
        \toprule
		Models & BLEU (\%) & Coverage (\%) & Length & Distinct-4 (\%) & Repetition-4 (\%)\\
        \midrule
	    Checklist & \bf 4.17 & 84.52** & 83.61** & 21.95** & 46.40**\\
        CVAE & 4.02 & 77.65** & 80.96** & 41.69** & 36.58**\\
        Pointer-S2S & 3.88** & 85.97** & 74.88** & 18.16** & 36.78**\\
        Link-S2S & 3.90** & 70.49** & \bf 95.65 & 16.64** & 59.83**\\
        \midrule
        PHVM (ours) & 2.85** & \bf 87.05 & 89.20** & \bf 72.87 & \bf 3.90\\
        w/o $z^p$ & 3.07** & 84.74** & 91.97** & 70.51** & 4.19\\
        w/o $z_t^s$ & 3.38** & 84.89** & 75.28** & 42.32** & 20.88**\\ 
	   \bottomrule
    \end{tabular}
    \caption{Automatic evaluation for advertising text generation. We applied bootstrap resampling~\cite{DBLP:conf/emnlp/Koehn04} for significance test. Scores that are significantly worse than the best results (in bold) are marked with ** for p-value $<$ 0.01. }
    \label{tab:res}
\end{table*}

Our dataset consists of three categories of clothing: tops, dress / skirt, and pants, which are further divided into 22, 23, and 9 types respectively (E.g., shirt, sweater are two types of tops).
Other categories (e.g., hats and socks) are discarded because these categories have insufficient data for training. The average length of advertising text is about 110 words.
To evaluate the expression diversity of our dataset, we computed distinct-4 (see Section 5.3) on 3,000 randomly sampled texts from our dataset. The distinct-4 score is 85.35\%, much higher than those of WIKIBIO~\cite{DBLP:conf/emnlp/LebretGA16} and ROTOWIRE~\cite{DBLP:conf/emnlp/WisemanSR17} (two popular data-to-text datasets). Therefore, our dataset is suitable for evaluating long and diverse text generation\footnote{We presented a detailed comparison with other benchmark corpora in Appendix \ref{appendix:cmp}.}. 

We left 1,070 / 3,127 instances for validation / test, and used the remainder for training.

\paragraph{Recipe Text Generation}
We used the same train-validation-test split (82,590 / 1,000 / 1,000) and pre-processing from \cite{DBLP:conf/emnlp/KiddonZC16}. In the training set, the average recipe length is 102 tokens, and the vocabulary size of recipe title / text is 3,793 / 14,103 respectively. The recipe dataset covers a wide variety of recipe types indicated by the vocabulary size of recipe title.

\subsection{Baselines}
We compared our model with four strong baselines where the former two do not perform planning and the latter two do:

\noindent \textbf{Checklist: }
This model utilizes an attention-based checklist mechanism to record what input data has been mentioned, and focuses more on what has not during generation~\cite{DBLP:conf/emnlp/KiddonZC16}.

\noindent \textbf{CVAE: }
The CVAE model proposed by \newcite{DBLP:conf/acl/ZhaoZE17} uses a latent variable to capture the diversity of responses in dialogue generation. We adapted it to our task by replacing the hierarchical encoder with a one-layer bidirectional GRU. 

\noindent \textbf{Pointer-S2S: }
A two-stage method~\cite{DBLP:journals/corr/abs-1809-00582} that decides the order of input data with Pointer Network~\cite{DBLP:conf/nips/VinyalsFJ15} before generation with Sequence-to-Sequence (Seq2Seq)~\cite{DBLP:journals/corr/BahdanauCB14}.

\noindent \textbf{Link-S2S}
Link-S2S~\cite{DBLP:conf/aaai/ShaMLPLCS18} is a Seq2Seq with link-based attention mechanism where a link matrix parameterizes the probability of describing one type of input item after another.

\subsection{Implementation Details}
For both advertising text generation and recipe text generation, the settings of our model have many in common. The dimension of word embedding is 300. All embeddings were randomly initialized. We utilized GRU for all RNNs. All RNNs, except the input encoder, the plan decoder, and the plan encoder, have a hidden size of 300. The global planning latent variable and local latent variables have 200 dimensions. We set batch size to 32 and trained our model using the Adam optimizer \cite{DBLP:journals/corr/KingmaB14} with a learning rate of 0.001 and gradient clipping threshold at 5. We selected the best model in terms of $\mathcal{L}_1+\mathcal{L}_2$ on the validation set.

As we need to train the plan decoder with full supervision, we extracted plans from the texts by recognizing attribute values (or ingredients) in each sentence with string match heuristics. Some sentences do not mention any input items; we associated these sentences with a special tag, which is treated as a special input item for Pointer-S2S, Link-S2S, and our model. Although our extraction method can introduce errors, the extracted plans are sufficient to train a good plan decoder\footnote{Our corpus and code are available at {\textit https://github.com/ZhihongShao/Planning-based-Hierarchical-Variational-Model.git}.}.

\noindent \textbf{Advertising Text Generation}
We embedded an attribute-value pair by concatenating the embedding of attribute and the embedding of attribute value. Embedding dimensions for attribute and attribute value are 30 and 100 respectively. The input encoder, the plan decoder, and the plan encoder all have a hidden size of 100.

\noindent \textbf{Recipe Text Generation}
We embedded a multi-word title (ingredient) by taking average pooling of the embeddings of its constituent words. Embedding dimensions for title word and ingredient word are 100 and 200 respectively. The input encoder, the plan decoder, and the plan encoder all have a hidden size of 200.

\begin{table*}[!t]
    \centering
	\small
    \begin{tabular}{*{9}{c}}
        \toprule
        \multirow{2}{*}{Models} & \multicolumn{3}{c}{Grammaticality} & \multirow{2}{*}{$\kappa$} & \multicolumn{3}{c}{Coherence} & \multirow{2}{*}{$\kappa$}\\
        \cmidrule(lr){2-4} \cmidrule(lr){6-8} & Win (\%) & Lose (\%) & Tie (\%) & & Win (\%) & Lose (\%) & Tie (\%)\\
        \midrule
        PHVM vs. Checklist & 59.0** & 23.5 & 17.5 & 0.484 & 54.5* & 42.5 & 3.0 & 0.425\\
        PHVM vs. CVAE & 69.5** & 13.5 & 17.0 & 0.534 & 60.0** & 37.0 & 3.0 & 0.426\\
        PHVM vs. Pointer-S2S & 76.5** & 17.0 & 6.5 & 0.544 & 56.5** & 39.0 & 4.5 & 0.414\\
        PHVM vs. Link-S2S & 66.0** & 28.5 & 5.5 & 0.462 & 62.5** & 31.5 & 6.0 & 0.415\\
        \bottomrule
    \end{tabular}
    \caption{Manual pair-wise evaluation for advertising text generation. We conducted Sign Test for significance test. Scores marked with * mean p-value $<$ 0.05 and ** for p-value $<$ 0.01. $\kappa$ denotes Fleiss' kappa, all indicating \textit{moderate agreement}.}
    \label{tab:cmp}
\end{table*}

\subsection{Automatic Evaluation Metrics}
We adopted the following automatic metrics. 
(1) \textbf{Corpus BLEU}: BLEU-4~\cite{DBLP:conf/acl/PapineniRWZ02}. 
(2) \textbf{Coverage}: This metric measures the average proportion of input items that are covered by a generated text. We recognized attribute values (ingredients) with string match heuristics. For the advertising text generation task, synonyms were also considered. 
(3) \textbf{Length}: The average length of the generated texts.
(4) \textbf{Distinct-4}: Distinct-n~\cite{DBLP:conf/naacl/LiGBGD16} is a common metric for diversity which measures the ratio of distinct n-grams in generated tokens. We adopted distinct-4. 
(5) \textbf{Repetition-4}: This metric measures redundancy with the percentage of generated texts that repeat at least one 4-gram.

\subsection{Advertising Text Generation}

\subsubsection{Automatic Evaluation}
Table \ref{tab:res} shows the experimental results. 
As our dataset possesses high expression diversity, there are many potential expressions for the same content, which leads to the low BLEU scores of all models.
Our model outperforms the baselines in terms of coverage, indicating that it learns to arrange more input items in a long text. With content ordering, Pointer-S2S outperforms both Checklist and CVAE in coverage. By contrast, our planning mechanism is even more effective in controlling generation: each sentence generation subtask is specific and focused, and manages to cover 95.16\% of the corresponding group on average. Noticeably, Link-S2S also models planning but has the lowest coverage, possibly because a static link matrix is unable to model flexible content arrangement in long text generation.
As for diversity, our model has substantially lower repetition-4 and higher distinct-4, indicating that our generated texts are much less redundant and more diversified. Notably, Link-S2S has the longest texts but with the highest repetition-4, which produces many redundant expressions.

To investigate the influence of each component in the hierarchical latent structure, we conducted ablation tests which removed either global latent variable $z^p$ or local latent variables $z_t^s$. As observed, removing $z^p$ leads to significantly lower distinct-4, indicating that $z^p$ contributes to expression diversity. The lower coverage is because the percentage of input items covered by a planning result decreases from 98.4\% to 94.4\% on average, which indicates that $z^p$ encodes useful information for planning completeness. When removing $z_t^s$, distinct-4 drops substantially, as the model tends to generate shorter and more common sentences. This indicates that $z_t^s$ contributes more to capturing variations of texts. The significantly higher repetition-4 is because removing $z_t^s$ weakens the dependencies among sentences so that the word decoder is less aware of the preceding generated context. The lower coverage is because each generated sentence covers less planned items (from 95.16\% to 93.07\% on average), indicating that $z_t^s$ keeps sentence $s_t$ more focused on its group. 

\begin{table*}[!t]
    \centering
	\small
    \begin{tabular}{*{6}{c}}
        \toprule
		Models & BLEU (\%) & Coverage (\%) & Length & Distinct-4 (\%) & Repetition-4 (\%)\\
        \midrule
	    Checklist $\S$ & 3.0 & 67.9 & N/A & N/A & N/A \\
	    Checklist & 2.6** & 66.9* & 67.59 & 30.67** & 39.1** \\
	    CVAE & 4.6 & 63.0** & 57.49** & 52.53** & 38.7**\\
        Pointer-S2S & 4.3 & 70.4** & 59.18** & 30.72** & 36.4** \\
        Link-S2S & 1.9** & 53.8** & 40.34** & 24.93** & 31.6**\\
        \midrule
        PHVM (ours) & \bf 4.6 & \bf 73.2 & \bf 70.92 & \bf 67.86 & \bf 17.3 \\
        \bottomrule
    \end{tabular}
    \caption{Automatic evaluation for recipe text generation. 
    Checklist was trained with its own source code. We also re-printed results from \cite{DBLP:conf/emnlp/KiddonZC16} (i.e., Checklist $\S$). We applied bootstrap resampling~\cite{DBLP:conf/emnlp/Koehn04} for significance test. Scores that are significantly worse than the best results (in bold) are marked with * for p-value $<$ 0.05 or ** for p-value $<$ 0.01. }
    \label{tab:res_recipe}
\end{table*}

\subsubsection{Manual Evaluation}
To better evaluate the quality of the generated texts, we conducted pair-wise comparisons manually. Each model generates texts for 200 randomly sampled inputs from the test set. We hired five annotators to give preference (win, lose or tie) to each pair of texts (ours vs. a baseline, 800 pairs in total).

\noindent \textbf{Metrics }
Two metrics were independently evaluated during annotation: \textbf{grammaticality} which measures whether a text is fluent and grammatical, and \textbf{coherence} which measures whether a text is closely relevant to input, logically coherent, and well-organized.

\noindent \textbf{Results }
The annotation results in Table \ref{tab:cmp} show that our model significantly outperforms baselines in both metrics. Our model produces more logically coherent and well-organized texts, which indicates the effectiveness of the planning mechanism. It is also worth noting that our model performs better in terms of grammaticality. The reason is that long text generation is decomposed into sentence generation sub-tasks which are easier to control, and our model captures inter-sentence dependencies through modeling the dependencies among local latent variables. 


\subsubsection{Diversity of Planning}
\begin{figure}[htb!]
    \includegraphics[width=0.5\textwidth]{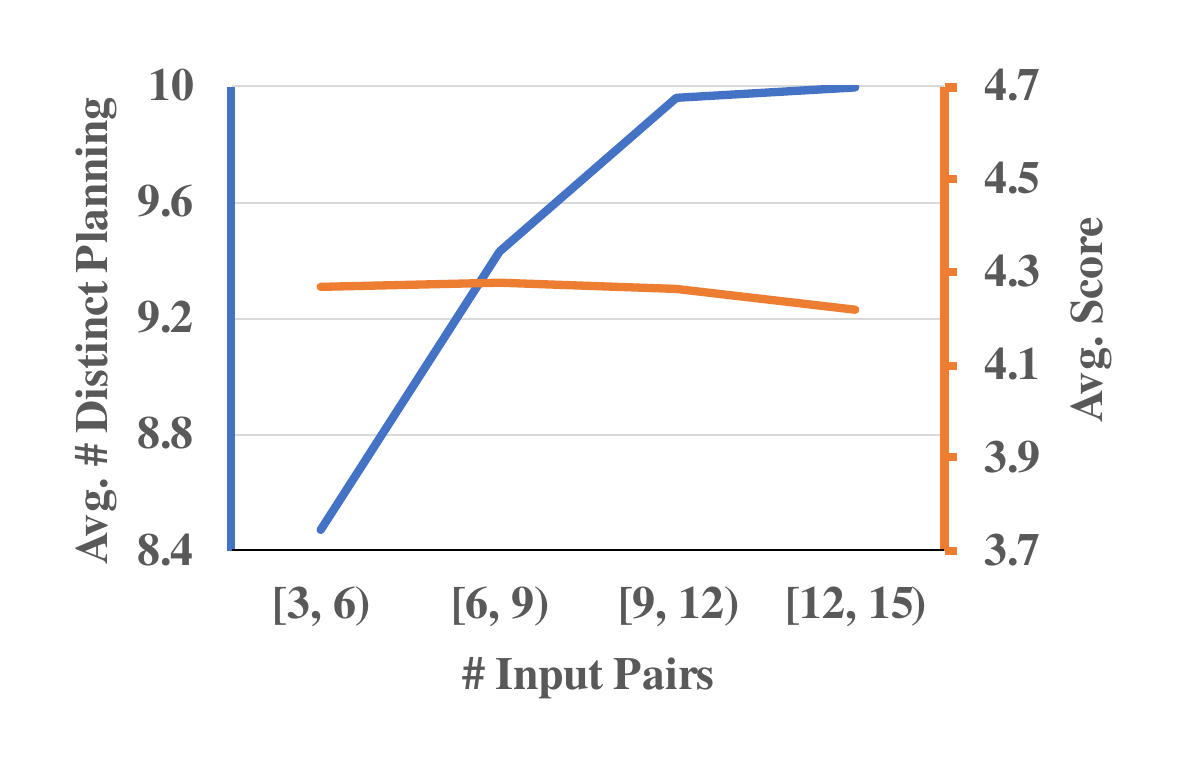}
    \caption{Average number of distinct planning results (left) / average score of generation quality (right) when the number of input pairs varies. }
    \label{fig:diversity}
\end{figure}
To evaluate how well our model can capture the diversity of planning, we conducted another manual evaluation. We randomly sampled 100 test inputs and generated 10 texts for each input by repeatedly sampling latent variables. Five annotators were hired to score (a Likert scale $\in [1, 5]$) a text about whether it is a qualified advertising text, which requires comprehensive assessment in terms of fluency, redundancy, content organization, and coherence. We computed the average of five ratings as the final score of a generated text.

\noindent \textbf{Results}
The average score of a generated text is 4.27. Among the 1,000 generated texts, 79.0\% of texts have scores above 4. These results demonstrate that our model is able to generate multiple high-quality advertising texts for the same input. 

We further analyzed how our model performs with different numbers of input attribute-value pairs (see Figure \ref{fig:diversity}). 
A larger number of input items indicates more potential reasonable ways of content arrangement. 
As the number of input items increases, our model produces more distinct planning results while still obtaining high scores (above 4.2). It indicates that our model captures the diversity of reasonable planning. The average score drops slightly when the number of input pairs is more than 12. This is due to insufficient training data for this range of input length (accounting for 6.5\% of the entire training set). 

To further verify the planning diversity, we also computed self-BLEU~\cite{DBLP:conf/sigir/ZhuLZGZWY18} to evaluate how different planning results (or texts) for the same input overlap (by taking one planning result (or text) as hypothesis and the rest 9 for the same input as reference and then computing BLEU-4). 
The average self-BLEU of the planning results is 43.37\% and that of the texts is 16.87\%, which demonstrates the much difference among the 10 results for the same input.

\noindent \textbf{Annotation Statistics }
The Fleiss' kappa is 0.483, indicating \textit{moderate agreement}.

\subsection{Recipe Text Generation}
Table \ref{tab:res_recipe} shows the experimental results. Our model outperforms baselines in terms of coverage and diversity; it manages to use more given ingredients and generates more diversified cooking steps. We also found that Checklist / Link-S2S produces the general phrase ``all ingredients" in 14.9\% / 24.5\% of the generated recipes, while CVAE / Pointer-S2S / PHVM produce the phrase in 7.8\% / 6.3\% / 5.0\% of recipes respectively. These results demonstrate that our model may generalize well to other data-to-text  generation tasks.

\section{Case Study}
We present examples for both tasks in Appendix \ref{appendix:case}.

\section{Conclusion and Future Work}
We present the Planning-based Hierarchical Variational Model (PHVM) for long and diverse text generation. A novel planning mechanism is proposed to better model input data and address the inter-sentence incoherence problem. PHVM also leverages a hierarchical latent structure to capture the diversity of reasonable planning and sentence realization. Experiments on two data-to-text corpora show that our model is more competitive to generate long and diverse texts than state-of-the-art baselines. 

Our planning-based model may be inspiring to other long text generation tasks such as long text machine translation and story generation. 


\section*{Acknowledgements}
This work was supported by the National Science Foundation of China (Grant No. 61936010/61876096) and the National Key R\&D Program of China (Grant No. 2018YFC0830200). We would like to thank THUNUS NExT Joint-Lab for the support.

\bibliography{emnlp-ijcnlp-2019}
\bibliographystyle{acl_natbib}
\clearpage
\appendix

\section{Dataset}

\subsection{Dataset Statistics}
\label{appendix:stat}
\begin{figure}[htb!]
    \includegraphics[width=0.5\textwidth]{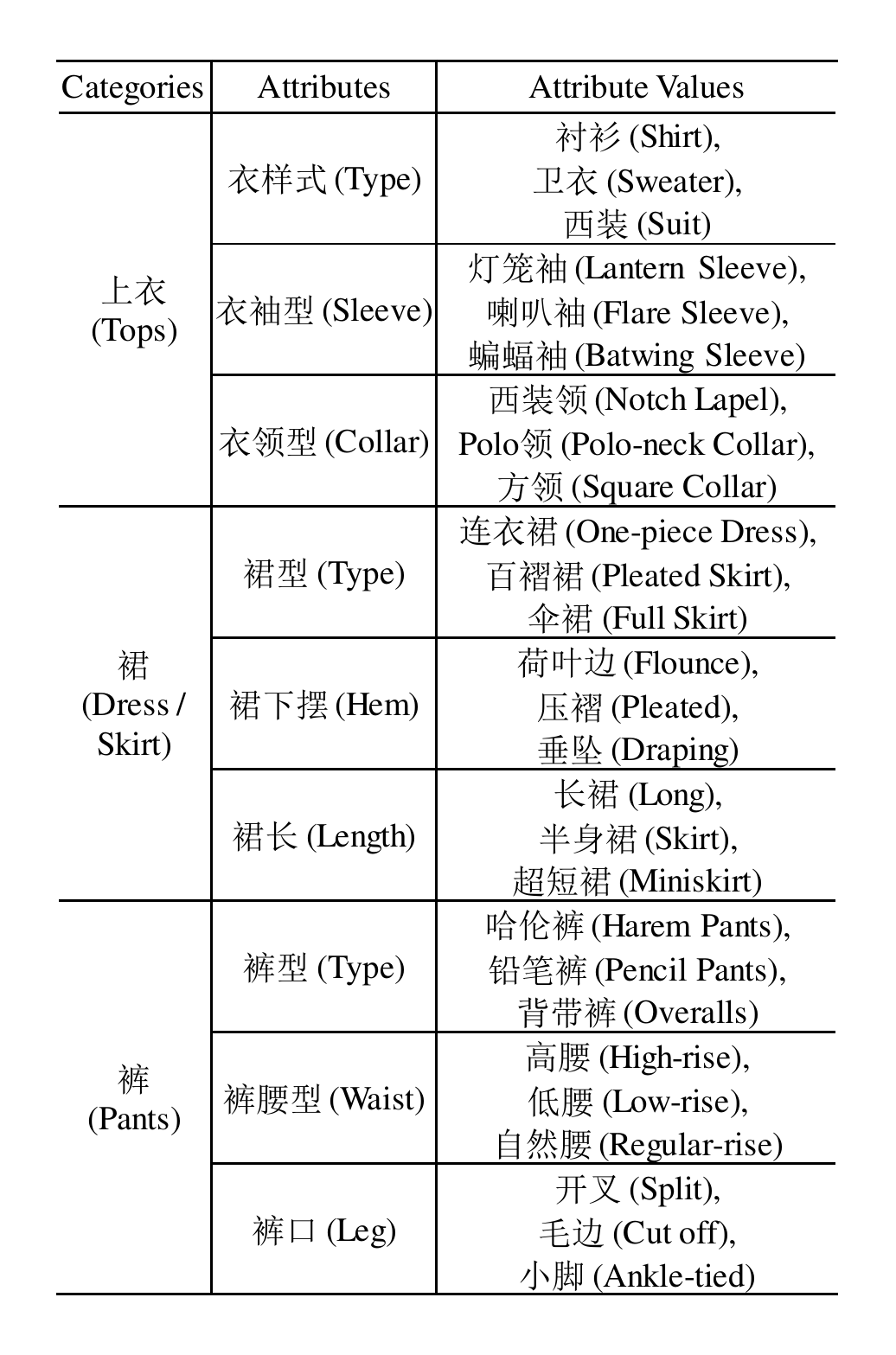}
    \caption{Samples of attributes and attribute values for tops, dress / skirt, and pants.}
    \label{fig:att}
\end{figure}
Figure \ref{fig:att} shows some samples of attributes and attribute values from our dataset for advertising text generation.

\subsection{Comparisons with other Benchmark Datasets}
\label{appendix:cmp}
\begin{table*}[htbp!]
\centering
\small
\begin{tabular}{*{7}{c}}
  \toprule
  & E2E & WebNLG & WG & WB & RW & Ours\\
  \midrule
  Avg. Len. & 14.3 & 22.69 & 28.7 & 26.1 & 337.1 & 110.2\\
  Distinct-4 & 35.44\% & 46.84\% & N/A & 47.04\% & 48.12\% & 85.35\% \\
  Vocab & 65.7K & 8K & 394 & 400K & 11.3K & 54.9K\\
  \# Instances & 50.6K & 25.3K & 22.1K & 728K & 4.9K & 119K\\
  \bottomrule
\end{tabular} 
\caption{\label{tab:cmp_data}Statistics of E2E, WebNLG, WEATHERGOV (WG), WIKIBIO (WB), ROTOWIRE (RW) and our dataset. We computed distinct-4 (see section 5.3) on 3,000 randomly sampled advertising texts for our dataset and on samples with comparable number of words for E2E, WebNLG, WB and RW, respectively.}
\end{table*}

Compared with other widely used data-to-text datasets (i.e., WEATHERGOV, WIKIBIO, E2E, WebBLG, and ROTOWIRE), our dataset is more suitable for long and diverse text generation. Table \ref{tab:cmp_data} presents the comparisons among the datasets. WEATHERGOV~\cite{DBLP:conf/acl/LiangJK09} defines a task to generate a weather forecast from weather statistics; it has a vocabulary size of less than 400, which indicates the simplicity of its language expressions. WIKIBIO~\cite{DBLP:conf/emnlp/LebretGA16} defines a task that requires to generate a very short biography (about 26 words) from personal information. E2E~\cite{DBLP:conf/sigdial/NovikovaDR17} and WebNLG~\cite{DBLP:conf/acl/GardentSNP17} are not confined to one specific domain, but also consist of very short texts (one sentence for most of the time). These datasets are not suitable for long text generation.  ROTOWIRE~\cite{DBLP:conf/emnlp/WisemanSR17} defines a task that requires to generate a summary from the records of a basketball game; the texts in ROTOWIRE are long enough but lack language expression diversity. We evaluated expression diversity with distinct-4. Before computing distinct-4, we first replaced numbers and named entities with their categories because such words contribute much to distinct-4 but not to expression diversity. For example, \textit{Lakers} was replaced with the tag \textit{$<$TEAM$>$} in ROTOWIRE, and \textit{Dior} was replaced with the tag \textit{$<$BRAND$>$} in our dataset. We then computed distinct-4 on 3,000 randomly sampled texts from our dataset and on samples with comparable number of words from E2E, WebNLG, WIKIBIO, and ROTOWIRE respectively. Our dataset exhibits much higher diversity with a substantially higher distinct-4 score than other corpora. As the texts in our dataset are long (the average length is about 110 words) and diverse, our dataset is suitable to evaluate long and diverse text generation in this paper.

Some other datasets pair structured data with user-generated reviews, such as Amazon reviews~\cite{DBLP:conf/sigir/McAuleyTSH15}, Yelp dataset\footnote{https://www.yelp.com/dataset/challenge}, and IMDb dataset~\cite{DBLP:conf/acl/MaasDPHNP11}. 
We did not use such corpora because the contents of user-generated reviews do not mainly come from the data but commonly depend on many other things such as the reviewers' experience and preference. 

\section{Case Study}
\label{appendix:case}
\subsection{Advertising Text Generation}
\begin{figure*}[t!]
    \includegraphics[width=\textwidth]{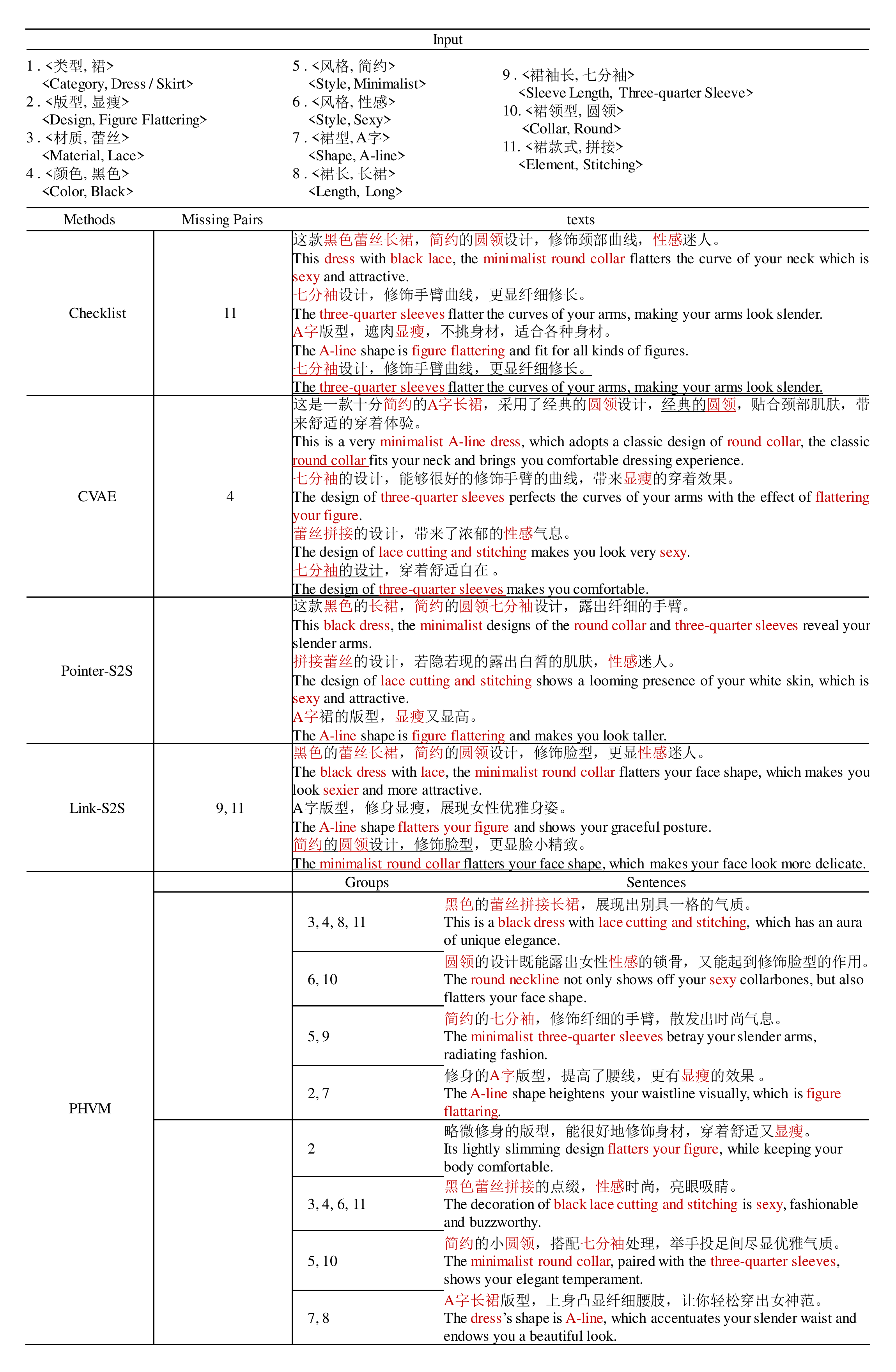}
    \caption{Generated advertising texts from different models. Attribute values are colored in red. Repeated expressions are underlined.}
    \label{fig:adv_case}
\end{figure*}

Figure \ref{fig:adv_case} shows generated texts from different models given the same input. 

Most baselines fail to cover all the provided data and repeatedly describe some of the input items. For example, the text from Link-S2S ignores the attribute value \textit{three-quarter sleeve} and describes the \textit{round collar} twice. Checklist and CVAE also have similar problems. As Link-S2S and Checklist inject variations only at the conditional output distribution, they suffer from the redundancy problem. Though Pointer-S2S covers all attribute values without redundancy, it introduces logical incoherence (the \textit{round collar} can not \textit{reveal slender arms}) in the first sentence. By contrast, both texts generated by our model cover all the input data without redundancy. 

Due to diverse yet reasonable planning, the two texts of our model exhibit different discourse structures. The first text adopts a general-to-specific discourse structure
where the statement in the beginning (i.e., {\it the elegance of the dress}) is supported by the following descriptions of local features.
It groups global features (i.e., \textit{color}, \textit{material} and \textit{length}) from the input in the first sentence and realizes each of the remaining sentences with one local feature. 
The second text adopts a parallel structure 
which splits global features and arranges some of them in the middle. 
Despite the difference, the two texts exhibit a global pattern in the data. They both describe the dress from top to bottom (i.e., \textit{collar} $->$ \textit{sleeve} $->$ \textit{shape of the lower part}), which verifies the effectiveness of content organization. Noticeably, the two texts show diverse wording, which exemplifies that our model captures the diversity of expressions.

\subsection{Recipe Text Generation}
\begin{figure*}[t!]
    \includegraphics[width=\textwidth]{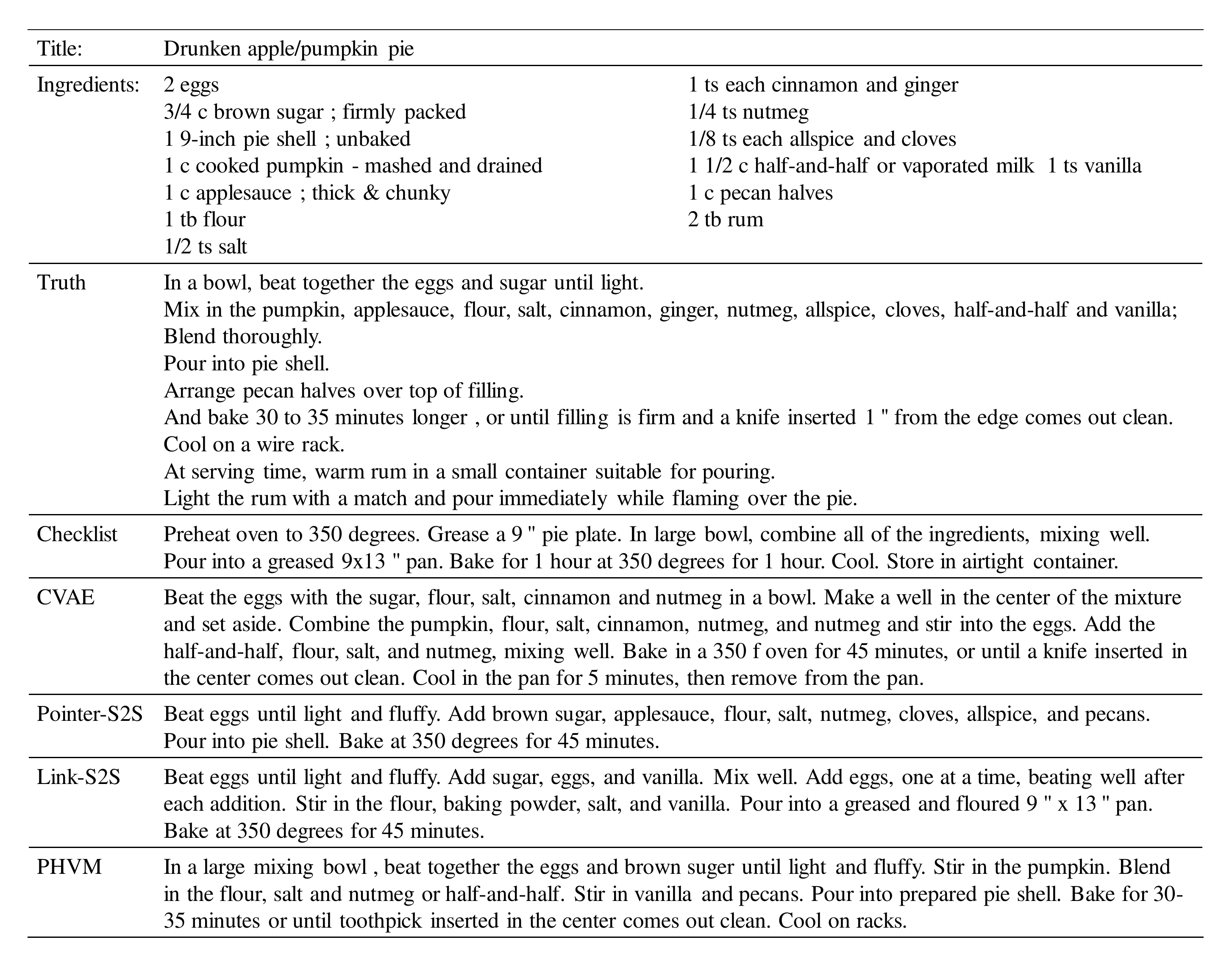}
    \caption{Generated recipes from different models.}
    \label{fig:recipe_case}
\end{figure*}
Figure \ref{fig:recipe_case} shows the generated examples. Although the three models fail to cover all given ingredients, our model gives the most complete procedure for making a \textit{pumpkin pie} which includes five steps: {\it 1.beat eggs $->$ 2.blend with some other ingredients $->$ 3.pour into pie shell $->$ 4.bake $->$ 5.cool}. Our model also gives the most specific and precise instructions for step 4 and step 5. By contrast, all baselines miss step 3 or step 5. Checklist produces the general phrase ``combine all of the ingredients". CVAE suffers from the redundancy and incoherence problems. Pointer-S2S mentions the most ingredients but misses the most important one ``pumpkin". Link-S2S misses ``pumpkin" and generates incoherent expressions.

\end{document}